\documentclass[letterpaper]{article} 
\usepackage[preprint]{aaai2027}
\usepackage[hyphens]{url}  
\usepackage{graphicx} 
\usepackage{natbib}  
\usepackage{caption} 
\usepackage{amsmath}
\usepackage{amssymb}
\usepackage{algorithm}
\usepackage{algorithmic}
\usepackage{booktabs}
\usepackage{multirow}
\usepackage{tabularx}
\definecolor{cvpr_blue}{cmyk}{0.65,0.35,0,0.12}

\copyrighttext{This manuscript is an arXiv preprint.}

\title{Breaking the Curse of Multilinguality in Many-to-Many Speech-to-Text Translation via a Resource-Aware Mixture of Speech Encoders}

\author{
Yexing Du\textsuperscript{\rm 1,\rm 2},
Kaiyuan Liu\textsuperscript{\rm 1,\rm 2},
Youcheng Pan\textsuperscript{\rm 2},
Bo Yang\textsuperscript{\rm 2},
Chengpeng Fu\textsuperscript{\rm 1,\rm 2},
Yu Wang\textsuperscript{\rm 2},
Ming Liu\textsuperscript{\rm 1,\rm 2}
}
\affiliations{
\textsuperscript{\rm 1}Harbin Institute of Technology, Harbin, China\\
\textsuperscript{\rm 2}Pengcheng Laboratory, Shenzhen, China\\
\{yxdu, mliu\}@ir.hit.edu.cn
}

\begin{document}

\maketitle

\begin{abstract}
Multimodal large language models (MLLMs) have achieved significant success in speech-to-text translation (S2TT). However, when processing multilingual speech inputs, a single speech encoder shared across all languages suffers from the curse of multilinguality: languages at different resource levels compete for limited representation capacity, leading to strong high-resource performance but substantial degradation on low-resource speech. To address this problem and improve multilingual consistency, we propose MSRT, a novel framework built around a resource-aware Mixture of Speech Encoders (MoSE). MoSE uses an explicit language router to assign each utterance to an appropriate expert encoder. A frozen expert preserves high-resource language capabilities, while a trainable expert adapts to and specializes in medium- and low-resource languages. We further introduce a five-stage curriculum learning strategy that substantially reduces data dependence, requiring only 10 hours of paired S2TT data per language for effective alignment. We conduct extensive experiments on 45 languages, systematically evaluating all $45 \times 44$ translation directions. Our 4B-parameter model achieves state-of-the-art performance, outperforming substantially larger baselines. Empirical analyses show that MoSE improves high-, medium-, and low-resource languages simultaneously, with the largest gains on low-resource speech, thereby breaking the curse of multilinguality without compromising high-resource performance. To support future multilingual S2TT research, we release our code and models.
\end{abstract}

\begin{links}
\link{Code}{https://github.com/yxduir/MSRT}
 \end{links}

\section{Introduction}

Speech-to-text translation (S2TT) aims to translate speech from a source language into text in a target language. 
Traditionally, these systems have utilized cascaded pipelines: an Automatic Speech Recognition (ASR) component first extracts a transcript from the speech~\cite{Radford2023}, which a subsequent Machine Translation (MT) model then converts into the desired language~\cite{shang2026scalingmodeldatamultilingual}.
Recently, there has been a growing focus on integrating speech encoders with large language models (LLMs) to endow them with multilingual speech capabilities, thereby advancing ASR~\cite{11346946} and S2TT tasks~\cite{du2026mcat}. 

\newpage
\begin{figure}[t]
    \centering
    \includegraphics[width=1\columnwidth]{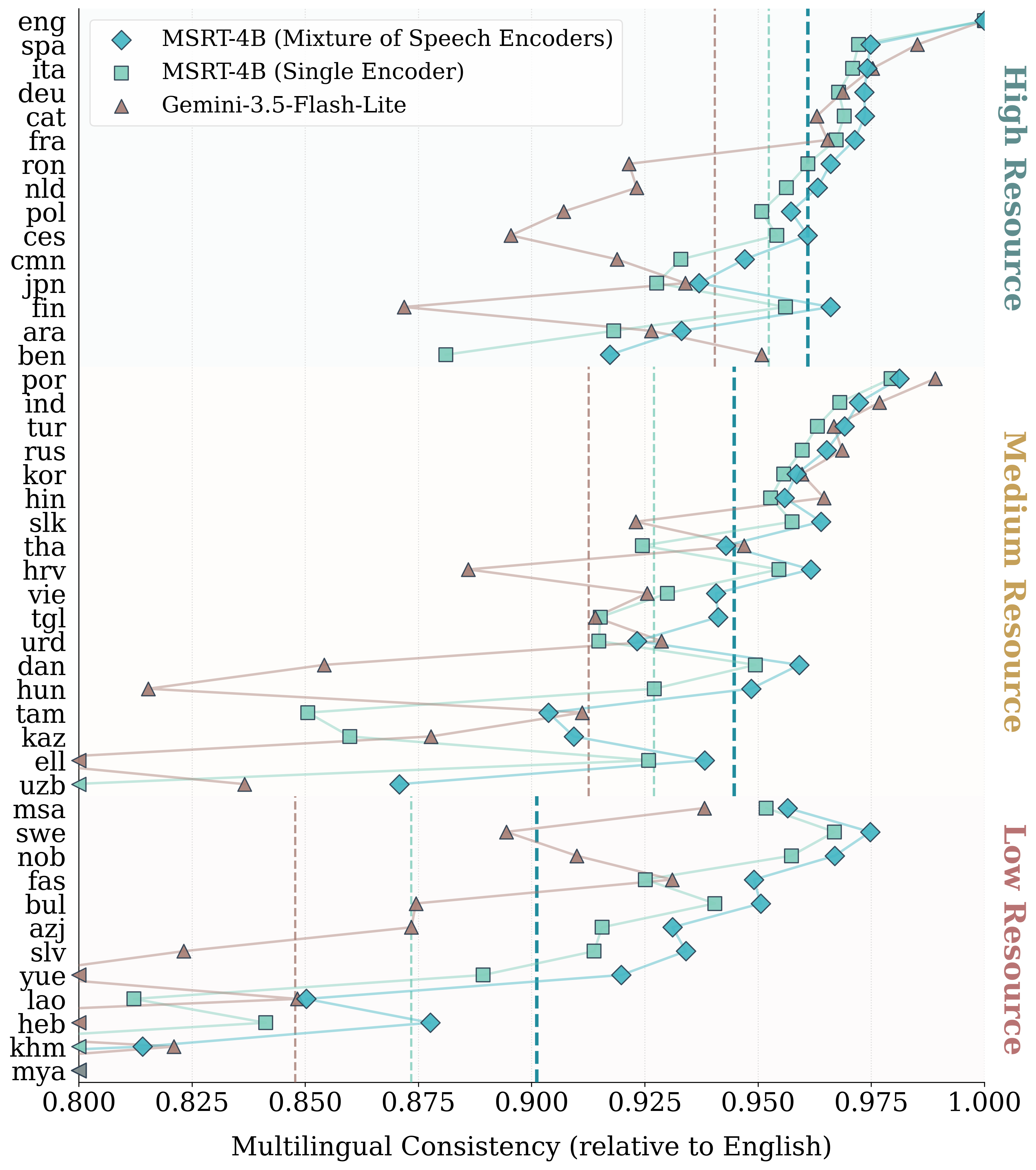}
\caption{Multilingual consistency by source language and resource level. Consistency is the ratio of a source language's average translation score (across 44 target directions) to the English average. Higher is better, and dashed vertical lines indicate each model's average within a resource group.}
    \label{fig:multilingual_consistency}
\end{figure}

Despite these advances, current MLLMs remain strongly English-centric in S2TT. They perform well on English-to-X directions but degrade in many-to-many translation from non-English speech. As shown in Figure~\ref{fig:multilingual_consistency}, their consistency scores for low-resource speech inputs fall substantially below those for English, revealing a pronounced cross-lingual performance gap. This disparity is not merely a target-generation issue: changing the source speech language causes substantial quality variation even when the target set and decoder remain unchanged. Consequently, English-centric results can overstate multilingual coverage and obscure failures on multilingual speech inputs.

One important cause of this inconsistency is that most multilingual MLLMs~\cite{xu2025qwen3} rely on a single shared speech encoder for all input languages. As language coverage expands, this architecture suffers from the ``\textbf{curse of multilinguality}''~\cite{conneau2020unsupervised}: limited encoder capacity forces languages with different levels of available resources to compete for the same representation space. High-resource languages require preserving pretrained acoustic knowledge, whereas medium- and low-resource languages require greater adaptation. Joint optimization can therefore overwrite established representations or underfit scarce-language features.

Inspired by the Mixture-of-Experts architectures in LLMs~\cite{blevins-etal-2024-breaking}, we propose MSRT, an S2TT MLLM centered on a resource-aware \textbf{Mixture of Speech Encoders (MoSE}). Two homogeneous Whisper encoders~\cite{Radford2023} assume asymmetric roles: a frozen expert handles high-resource languages, while a trainable expert adapts to medium- and low-resource speech. A source-language token routes each utterance to one expert, and a target-language token controls generation. This design provides specialization at the inference cost of a single encoder. We further extend the curriculum strategy~\cite{du2025making} into five stages of increasing difficulty, progressively establishing speech--text alignment and translation.

We evaluate MSRT on FLEURS across all $45\times44$ non-identity translation directions, with detailed comparisons on representative languages from 11 language families. MSRT-4B achieves the highest average COMET score of 83.3, outperforming both the Gemini-3.5-Flash-Lite API and the substantially larger MCAT-27B~\cite{du2026mcat}. Across the complete direction grid, MSRT obtains COMET scores of at least 80 on 1,552 directions (78.4\%), compared with 928 for Gemini. Ablation studies show that MoSE improves performance across high-, medium-, and low-resource languages, with the largest gains on low-resource languages, while further strengthening high-resource performance. These consistent gains across all resource groups demonstrate that MoSE breaks the curse of multilinguality rather than trading high-resource performance for low-resource improvements. Experiments on CoVoST-2 further demonstrate robust cross-dataset generalization.

Our contributions are summarized as follows:
\begin{itemize}
    \item We propose MoSE, a resource-aware mixture-of-speech-encoder architecture that breaks the curse of multilinguality by preserving high-resource knowledge while enabling targeted adaptation for medium- and low-resource languages.
    \item We introduce MSRT-4B, a compact 4B-parameter model that supports all translation directions among 45 languages using only 10 hours of paired S2TT data per language. We open-source its code and model to facilitate reproducible research.
\item We comprehensively evaluate all 1,980 translation directions across 45 languages, demonstrating broad and reliable many-to-many translation across diverse language families and resource levels, especially for low-resource speech in realistic multilingual settings.
\end{itemize}

\section{Related Work}

\paragraph{Speech-to-Text Translation.}
Conventional speech-to-text translation (S2TT) cascades automatic speech recognition and machine translation. Although this design can reuse strong task-specific models, recognition errors propagate downstream and two-stage decoding increases latency. End-to-end S2TT instead learns a direct mapping from source speech to target text. SeamlessM4T integrates multilingual speech recognition and translation with text translation in a single model, demonstrating that one architecture can support diverse speech--text tasks and languages \cite{Barrault2023}. Its success, however, still depends heavily on paired supervision. ZeroSwot reduces this dependence by connecting an ASR-trained speech encoder to a massively multilingual translation model through CTC compression and optimal-transport alignment, enabling direct S2TT without paired translation data \cite{tsiamas2024pushing}. These studies establish speech--text alignment as a central challenge in end-to-end translation.

\paragraph{Many-to-Many Speech-to-Text Translation.}
Many-to-many S2TT expands both source- and target-language coverage, requiring one model to preserve language-specific acoustic cues while controlling multilingual generation. SeamlessM4T demonstrates broad multilingual coverage through shared recognition and translation components \cite{Barrault2023}. General-purpose multimodal LLMs such as Qwen3-Omni further show that pretrained language backbones can support multilingual speech understanding and translation \cite{xu2025qwen3}. Specialized MLLMs more directly exploit the translation knowledge of LLMs: LLM-SRT transfers it to many-to-many S2TT through curriculum learning \cite{du2025making}, while MCAT scales MLLM-based translation to 70 languages \cite{du2026mcat}. Nevertheless, increasing language coverage does not eliminate resource imbalance or interference between languages. This limitation motivates speech-side mechanisms that allocate adaptation capacity according to the data resources and acoustic variation of different languages.

\paragraph{Mixture of Experts.}
MoE has been explored in multimodal and speech models from several perspectives. Uni-MoE activates modality-specific experts to reduce cross-modal interference in a unified MLLM \cite{li_unimoe}, whereas SimulMega uses expert routers as policies for balancing latency and quality in simultaneous translation \cite{le2026simulmega}. PaM selects heterogeneous audio encoders from task prompts to serve different audio tasks \cite{shan-etal-2025-enhancing}, while DialectMoE learns dynamic expert routing for multi-dialect ASR \cite{jie-etal-2024-dialectmoe}. In contrast, this work explores MoE specifically for multilingual S2TT and focuses on interference induced by imbalanced language resources. MSRT routes source languages by resource level between two homogeneous Whisper encoders with asymmetric roles: a frozen expert preserves high-resource representations, and a trainable expert specializes in medium- and low-resource speech. Thus, its explicit language-level routing investigates multilingual expert specialization without token-level load balancing or task-dependent expert selection.

\begin{figure*}[!t]
 \centering 
    \includegraphics[width=\linewidth]{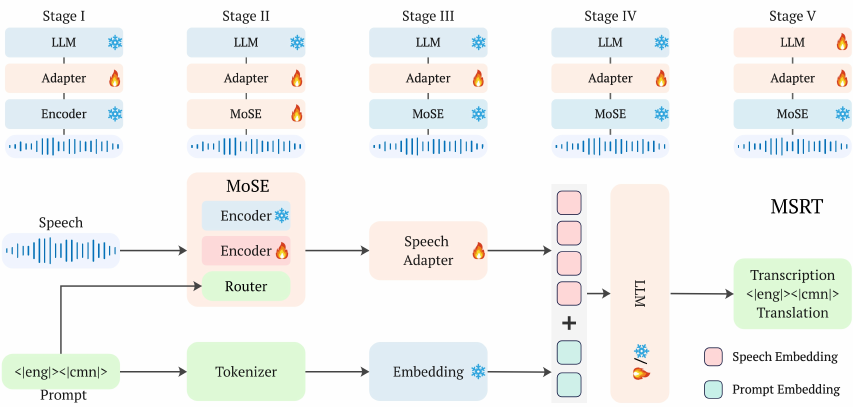}
\caption{Overview of the MSRT framework. Based on the source-language token in the prompt, MoSE routes high-resource speech to a frozen encoder and medium- and low-resource speech to a trainable encoder.}
    \label{fig:framework}
\end{figure*}

\section{Methodology}

\subsection{Model Architecture}
As shown in Figure~\ref{fig:framework}, MSRT connects a resource-aware Mixture of Speech Encoders (MoSE), a speech adapter, and a pretrained LLM. 

\paragraph{Mixture of Speech Encoders.}

MoSE comprises two encoders and an explicit language router. Instead of sharing one encoder across languages with unequal supervision, it separates high-resource knowledge preservation from medium- and low-resource adaptation while activating one expert per utterance. This design breaks the curse of multilinguality by improving underrepresented languages without sacrificing high-resource performance.

\begin{itemize}
    \item \textbf{Frozen Expert.} The frozen expert $E_f$ retains the original Whisper encoder with fixed parameters. It preserves general speech representations for high-resource languages, prevents catastrophic forgetting, and anchors the shared speech interface during adaptation.
    \item \textbf{Trainable Expert.} The trainable expert $E_t$ shares the Whisper initialization but adapts to medium- and low-resource speech, capturing underrepresented accents and phonological patterns. Its architecture matches $E_f$, so both experts produce compatible representations and share one adapter without expert-specific projections.
    \item \textbf{Explicit Language Router.} The router maps each source-language token to its resource-group expert: high-resource languages use $E_f$, whereas medium- and low-resource languages use $E_t$. Table~\ref{tab:language_support_summary} lists all assignments. Utterance-level routing keeps all frames on one acoustic path and requires neither a learned gate nor a balancing loss.
\end{itemize}

Overall, let $\mathcal{R}(\ell_s)$ denote the expert selected for source language $\ell_s$. The MoSE output is written as
\begin{equation}
H_{\mathrm{MoSE}}(x,\ell_s)
=E_{\mathcal{R}(\ell_s)}(x),\quad
\mathcal{R}(\ell_s)\in\{f,t\}.
\label{eq:mse_routing}
\end{equation}
Here, $\mathcal{R}(\ell_s)=f$ for $\ell_s\in\mathcal{L}_{\mathrm{high}}$ and $\mathcal{R}(\ell_s)=t$ otherwise, covering the medium- and low-resource groups. This formulation retains the inference cost of a single encoder.

\paragraph{Speech Adapter.}
The selected encoder produces variable-length features. A Q-Former compresses them into a translation-aware representation, while an MLP projects it into the LLM hidden space. Together, these modules remove acoustic redundancy while retaining information required for translation.

\paragraph{LLM.}
The pretrained LLM provides multilingual translation knowledge and remains frozen during speech-side alignment to prevent catastrophic forgetting. It is unlocked only in the final training stage, where it adapts to the aligned acoustic representation.

\paragraph{Tokenizer Expansion.}
We add dedicated language tokens for the FLEURS languages~\cite{Conneau2023}. The source-language token controls MoSE encoder selection, while the target-language token specifies the LLM translation direction. Together, they provide explicit language control and reduce unintended switching.

\paragraph{Text Prompt.}
As shown in Figure~\ref{fig:framework} and Table~\ref{tab:prompt_design}, the prompt explicitly instructs the LLM to translate from the source language into the target language. The source-language token also controls encoder selection through the explicit language router, while the target-language token specifies the desired output language.

\begin{table}[h]
\centering
\small
\caption{Prompt design for the five-stage curriculum.}
\label{tab:prompt_design}
\setlength{\tabcolsep}{1pt}
\begin{tabular}{clcll}
\toprule
\textbf{Stage} & \textbf{Task} & \textbf{Speech} & \textbf{Text Prompt} & \textbf{Prediction} \\
\midrule
I  &ASR & \textcolor{cvpr_blue}{\checkmark} & <|eng|> & Text \\
II &ASR & \textcolor{cvpr_blue}{\checkmark} & <|eng|> & Text \\
III &SMT & \textcolor{cvpr_blue}{\checkmark} & Text<|eng|><|cmn|> & MT \\
IV &SRT & \textcolor{cvpr_blue}{\checkmark} & <|eng|><|cmn|> & Text<|eng|><|cmn|> MT \\
V &SRT & \textcolor{cvpr_blue}{\checkmark} & <|eng|><|cmn|> & Text<|eng|><|cmn|> MT \\
\bottomrule
\end{tabular}
\end{table}

\subsection{Five-Stage Curriculum Learning}
Following \citet{du2025making}, we organize ASR, speech-guided machine translation (SMT), and joint speech recognition and translation (SRT) into the five stages in Table~\ref{tab:prompt_design}. Table~\ref{tab:parameters} summarizes which modules are optimized at each stage; every stage starts from the preceding checkpoint.
\begin{itemize}
    \item \textbf{Stage I: ASR alignment.} The Q-Former and MLP adapter are trained successively on English-only, high-resource multilingual, and fully multilingual ASR data, establishing a stable multilingual speech--text interface while both speech experts and the LLM remain frozen.
    \item \textbf{Stage II: Expert specialization.} MoSE routing is activated for multilingual ASR. The frozen expert $E_f$ preserves high-resource representations, whereas $E_t$ is optimized for medium- and low-resource speech; the shared adapter continues training.
    \item \textbf{Stage III: Translation activation.} SMT supplies the ground-truth transcription together with source- and target-language tokens. The adapter is optimized to connect routed speech features to the translation knowledge of the frozen LLM.
    \item \textbf{Stage IV: End-to-end SRT.} The ground-truth transcription is removed from the input. With the experts and LLM fixed, the adapter learns to generate both the transcription and translation directly from routed speech.
    \item \textbf{Stage V: Joint adaptation.} LLM LoRA is enabled and jointly optimized with the adapter for SRT, adapting multilingual generation to acoustic inputs without updating the pretrained LLM weights.
\end{itemize}

\begin{table}[h]
\centering
\small
\caption{MLLM settings for MSRT.}
\label{tab:parameters}
\renewcommand{\arraystretch}{1}
\setlength{\tabcolsep}{5pt}
\begin{tabular}{lcl}
\toprule
\textbf{Modules} & \textbf{Stage} & \textbf{Details} \\
\midrule
Speech Encoder (MoSE) & & \\
\hspace{1em}Frozen Expert ($E_f$) & -- & For high-resource \\
\hspace{1em}Trainable Expert ($E_t$) & II & For medium-/low-resource \\
\hspace{1em}Language Router & -- & Explicit expert routing \\
\midrule
Speech Adapter & & \\
\hspace{1em}Q-Former & All & Feature compression \\
\hspace{1em}MLP & All & Dimension alignment \\
\midrule
LLM & & \\
\hspace{1em}$+$ Vocabulary & -- & Language tokens \\
\hspace{1em}$+$ LoRA & V & $r=16$, $\alpha=32$ \\
\bottomrule
\end{tabular}
\end{table}

\section{Experimental Settings}

\paragraph{Datasets.}
We use {Common Voice 24}~\cite{ardila-etal-2020-common} and {FLEURS} for multilingual ASR pre-training, combining broad language coverage with curated speech data. For S2TT training, we use paired speech--text data from FLEURS, comprising approximately 10 hours of speech per language. We evaluate the S2TT models on both FLEURS and the {CoVoST-2}~\cite{Wang2021} benchmark.

\paragraph{Training Details.}
As shown in Table~\ref{tab:parameters}, the MLLM comprises a MiLMMT-4B LLM~\cite{shang2026scalingmodeldatamultilingual}, MoSE, and a speech adapter containing a Q-Former and an MLP. The Q-Former uses 80 learnable queries to compress variable-length speech into a fixed-length interface for the LLM. We train in BF16 with DeepSpeed ZeRO-0 and AdamW, using a peak learning rate of $1\times10^{-4}$ and 1,000 warmup steps. Training takes less than three days on eight Ascend 910C NPUs. The same implementation can run on eight NVIDIA A100 GPUs with comparable accuracy.

\paragraph{Baselines.}
We compare three representative deployment paradigms:
\begin{itemize}
    \item \textbf{API-based:} Gemini-3.5-Flash-Lite~\cite{team2023gemini} serves as the API baseline.
    \item \textbf{Cascade:} Whisper-Large-v3~\cite{Radford2023} first transcribes speech, after which NLLB-200-3.3B~\cite{nllb2024scaling} translates the transcription.
    \item \textbf{End-to-end:} SeamlessM4T-V2-Large~\cite{Barrault2023}, Qwen3-Omni-30B~\cite{xu2025qwen3}, and MCAT-27B~\cite{du2026mcat} directly generate translations from speech and cover both specialized and general-purpose multilingual architectures.
\end{itemize}

\paragraph{Evaluation Metrics.}
We employ COMET~\cite{Rei2022} and spBLEU~\cite{Post2018} as our evaluation metrics. Specifically, spBLEU uses the FLORES-200 tokenizer. All results use a beam size of 1 unless stated otherwise.

\begin{table}[h]
\centering
\small
\setlength{\tabcolsep}{1pt}
\caption{Language support. Resource levels are defined following SeamlessM4T~\cite{Barrault2023}.}
\label{tab:language_support_summary}
\begin{tabular}{@{}l@{\hspace{6pt}}p{0.76\columnwidth}@{}}
\toprule
\textbf{Resource} & \textbf{Languages (ISO Code)} \\
\midrule
High & Arabic (ara), Bengali (ben), Catalan (cat) \\
($E_f$)    & Czech (ces), Chinese (cmn), German (deu) \\
     & English (eng), Finnish (fin), French (fra) \\
     & Italian (ita), Japanese (jpn), Dutch (nld) \\
     & Polish (pol), Romanian (ron), Spanish (spa) \\
\midrule
Medium & Danish (dan), Greek (ell), Hindi (hin) \\
($E_t$)       & Croatian (hrv), Hungarian (hun), Indonesian (ind) \\
       & Kazakh (kaz), Korean (kor), Portuguese (por) \\
       & Russian (rus), Slovak (slk), Tamil (tam) \\
       & Tagalog (tgl), Thai (tha), Turkish (tur) \\
       & Urdu (urd), Uzbek (uzb), Vietnamese (vie) \\
Low    & Azerbaijani (azj), Bulgarian (bul), Persian (fas) \\
($E_t$)      & Hebrew (heb), Khmer (khm), Lao (lao) \\
       & Malay (msa), Burmese (mya), Norwegian (nob) \\
       & Slovenian (slv), Swedish (swe), Cantonese (yue) \\
\midrule
\textbf{Total} & \textbf{45 languages (15 high, 18 medium, and 12 low)} \\
\bottomrule
\multicolumn{2}{@{}l}{\footnotesize
\hspace*{0em}$E_f$ denotes the frozen expert, and $E_t$ denotes the trainable expert.
} \\
\end{tabular}
\end{table}

\begin{table*}[!t]
\centering
\small
\caption{spBLEU / COMET results for $11\times44$ and $44\times11$ directions on FLEURS.}
\label{comet_1144}
\begin{tabularx}{\textwidth}{l*{6}{>{\centering\arraybackslash}X}}
\toprule
\multirow{2}{*}{\textbf{Direction}} & Gemini-3.5 & Whisper+ & SeamlessM4T & Qwen3-Omni & MCAT & MSRT-4B \\
 & -Flash-Lite & NLLB-3.3B & -V2-Large & -30B-Instruct & -27B & (ours) \\
\midrule
ara$\rightarrow$44 & \underline{22.2} / \underline{80.2} & 21.1 / 78.7 & 15.7 / 70.1 & 20.5 / 78.8 & 20.0 / 79.7 & \textcolor{cvpr_blue}{\textbf{23.7}} / \textcolor{cvpr_blue}{\textbf{82.0}} \\
cmn$\rightarrow$44 & 17.2 / 79.5 & 18.5 / 80.5 & 13.4 / 74.0 & \textcolor{cvpr_blue}{\textbf{20.9}} / \underline{82.8} & 17.7 / 81.3 & \underline{20.8} / \textcolor{cvpr_blue}{\textbf{83.2}} \\
eng$\rightarrow$44 & \textcolor{cvpr_blue}{\textbf{34.2}} / 86.6 & 30.5 / 84.3 & 31.8 / 85.3 & 31.9 / 85.7 & 31.9 / \underline{87.1} & \textcolor{cvpr_blue}{\textbf{34.2}} / \textcolor{cvpr_blue}{\textbf{87.9}} \\
hun$\rightarrow$44 & 14.8 / 70.6 & \underline{21.2} / 79.9 & 15.3 / 68.9 & 11.0 / 65.0 & 19.6 / \underline{79.9} & \textcolor{cvpr_blue}{\textbf{23.8}} / \textcolor{cvpr_blue}{\textbf{83.4}} \\
ind$\rightarrow$44 & \textcolor{cvpr_blue}{\textbf{27.4}} / \underline{84.6} & 23.7 / 82.2 & 15.2 / 71.5 & 24.7 / 83.4 & 24.2 / 84.2 & \underline{27.3} / \textcolor{cvpr_blue}{\textbf{85.4}} \\
jpn$\rightarrow$44 & 18.4 / 80.8 & 18.9 / 80.5 & 11.9 / 69.5 & \underline{19.5} / \underline{81.8} & 18.0 / 80.8 & \textcolor{cvpr_blue}{\textbf{20.4}} / \textcolor{cvpr_blue}{\textbf{82.3}} \\
kor$\rightarrow$44 & 20.8 / 83.1 & 19.5 / 81.8 & 14.2 / 74.1 & 20.4 / 82.7 & \underline{20.9} / \underline{83.4} & \textcolor{cvpr_blue}{\textbf{22.4}} / \textcolor{cvpr_blue}{\textbf{84.2}} \\
tam$\rightarrow$44 & \underline{17.3} / \underline{78.9} & 12.9 / 71.4 & 12.6 / 69.8 & 2.5 / 50.4 & 13.0 / 74.0 & \textcolor{cvpr_blue}{\textbf{17.9}} / \textcolor{cvpr_blue}{\textbf{79.4}} \\
tha$\rightarrow$44 & \underline{20.2} / \underline{82.0} & 17.1 / 78.6 & 11.5 / 69.0 & 18.8 / 81.3 & 15.6 / 78.9 & \textcolor{cvpr_blue}{\textbf{20.8}} / \textcolor{cvpr_blue}{\textbf{82.9}} \\
tur$\rightarrow$44 & \underline{25.6} / 83.7 & 23.4 / 83.0 & 15.8 / 72.1 & 23.2 / 82.5 & 24.0 / \underline{84.1} & \textcolor{cvpr_blue}{\textbf{25.7}} / \textcolor{cvpr_blue}{\textbf{85.2}} \\
vie$\rightarrow$44 & 19.7 / 80.1 & 18.7 / 78.9 & 14.1 / 72.6 & \underline{20.0} / \underline{81.3} & 18.0 / 80.5 & \textcolor{cvpr_blue}{\textbf{21.4}} / \textcolor{cvpr_blue}{\textbf{82.7}} \\ \midrule
44$\rightarrow$cmn & 21.1 / 78.3 & 14.3 / 71.2 & 11.3 / 65.4 & 21.3 / 78.4 & \underline{21.5} / \underline{80.2} & \textcolor{cvpr_blue}{\textbf{25.2}} / \textcolor{cvpr_blue}{\textbf{83.3}} \\
44$\rightarrow$eng & 24.9 / 79.2 & \underline{31.7} / 80.5 & 31.1 / \underline{83.3} & 28.1 / 79.7 & 27.5 / 80.6 & \textcolor{cvpr_blue}{\textbf{34.0}} / \textcolor{cvpr_blue}{\textbf{84.1}} \\
44$\rightarrow$hun & \underline{20.6} / 79.7 & 18.4 / 75.7 & 13.8 / 67.9 & 15.9 / 74.4 & 17.8 / \underline{79.7} & \textcolor{cvpr_blue}{\textbf{21.4}} / \textcolor{cvpr_blue}{\textbf{82.4}} \\
44$\rightarrow$ind & 21.7 / 80.9 & \underline{22.9} / 80.5 & 16.1 / 76.7 & 21.0 / 79.0 & 21.9 / \underline{82.1} & \textcolor{cvpr_blue}{\textbf{25.3}} / \textcolor{cvpr_blue}{\textbf{85.0}} \\
44$\rightarrow$jpn & \underline{18.2} / 81.8 & 9.3 / 76.1 & 7.3 / 71.9 & 17.2 / 81.0 & 18.2 / \underline{83.1} & \textcolor{cvpr_blue}{\textbf{19.9}} / \textcolor{cvpr_blue}{\textbf{85.4}} \\
44$\rightarrow$kor & \underline{16.0} / 79.9 & 13.9 / 77.5 & 10.2 / 71.8 & 15.2 / 78.7 & 15.5 / \underline{80.7} & \textcolor{cvpr_blue}{\textbf{16.8}} / \textcolor{cvpr_blue}{\textbf{83.2}} \\
44$\rightarrow$tam & \underline{20.0} / 80.0 & 19.8 / 78.5 & 17.6 / 75.9 & 15.1 / 74.7 & 19.0 / \underline{81.0} & \textcolor{cvpr_blue}{\textbf{21.2}} / \textcolor{cvpr_blue}{\textbf{83.1}} \\
44$\rightarrow$tha & \underline{27.6} / 78.1 & 20.2 / 74.0 & 20.1 / 70.3 & 26.7 / 78.1 & 26.2 / \underline{79.8} & \textcolor{cvpr_blue}{\textbf{29.9}} / \textcolor{cvpr_blue}{\textbf{82.5}} \\
44$\rightarrow$tur & \underline{19.6} / 78.4 & 18.3 / 76.0 & 13.6 / 70.0 & 15.8 / 73.8 & 18.6 / \underline{78.9} & \textcolor{cvpr_blue}{\textbf{21.1}} / \textcolor{cvpr_blue}{\textbf{81.6}} \\
44$\rightarrow$vie & 23.5 / 79.4 & \underline{24.0} / 78.7 & 18.4 / 73.8 & 22.2 / 77.3 & 23.2 / \underline{81.1} & \textcolor{cvpr_blue}{\textbf{27.1}} / \textcolor{cvpr_blue}{\textbf{83.9}} \\
\midrule
\textbf{$11{\to}44$} & \underline{21.6} / 80.9 & 20.5 / 80.0 & 15.6 / 72.4 & 19.4 / 77.8 & 20.3 / \underline{81.3} & \textcolor{cvpr_blue}{\textbf{23.5}} / \textcolor{cvpr_blue}{\textbf{83.5}} \\
\textbf{$44{\to}11$} & \underline{21.1} / 79.1 & 19.4 / 76.8 & 16.0 / 72.6 & 19.8 / 77.2 & 21.0 / \underline{80.5} & \textcolor{cvpr_blue}{\textbf{24.1}} / \textcolor{cvpr_blue}{\textbf{83.2}} \\
\textbf{Avg.} & \underline{21.3} / 80.0 & 20.0 / 78.4 & 15.8 / 72.5 & 19.6 / 77.5 & 20.6 / \underline{80.9} & \textcolor{cvpr_blue}{\textbf{23.8}} / \textcolor{cvpr_blue}{\textbf{83.3}} \\
\bottomrule
\multicolumn{7}{@{}l}{\footnotesize
\hspace*{0.4em}\underline{Underlining} denotes the second-best results, while
\textcolor{cvpr_blue}{\textbf{bold blue}} values indicate the best performance.} \\
\end{tabularx}
\end{table*}

\section{Experiments}
\subsection{Main Results}

\paragraph{Language Selection.}
As shown in Table~\ref{comet_1144}, the 11 languages cover 11 language families: Arabic (Afro-Asiatic), Mandarin Chinese (Sino-Tibetan), English (Indo-European), Hungarian (Uralic), Indonesian (Austronesian), Japanese (Japonic), Korean (Koreanic), Tamil (Dravidian), Thai (Kra--Dai), Turkish (Turkic), and Vietnamese (Austroasiatic).

\paragraph{Overall Analysis.}
According to the overall average in Table~\ref{comet_1144}, MSRT-4B achieves the highest COMET score of 83.3, outperforming MCAT-27B (80.9), Gemini-3.5-Flash-Lite (80.0), Whisper+NLLB-200-3.3B (78.4), Qwen3-Omni (77.5), and SeamlessM4T-V2-Large (72.5). Moreover, MSRT-4B ranks first in both the $11\times44$ and $44\times11$ settings, showing that its overall advantage is sustained across translation directions rather than driven by performance in a single setting.

\paragraph{Translation Direction Analysis.}
Performance differs notably between $11\times44$ and $44\times11$ directions. The cascaded system achieves a competitive COMET score of 80.0 in the former setting but drops to 76.8 in the latter, indicating that translation from a broader set of source speech languages is more challenging, partly due to accumulated multilingual recognition errors. In contrast, MSRT-4B achieves COMET scores of 83.5 and 83.2, respectively, demonstrating the strongest performance and robust consistency across translation directions.

\subsection{Eng$\rightarrow$44 S2TT}

\paragraph{Overall Analysis of Eng$\rightarrow$44 Directions.}
Table~\ref{tab:eng44_comet} reports COMET performance across all 44 English-source translation directions, including the newly evaluated Gemini-3.5-Flash-Lite baseline. MSRT-4B achieves the highest average score of \textbf{87.9}, outperforming MCAT-27B, Gemini-3.5-Flash-Lite, Qwen3-Omni, and SeamlessM4T-V2-Large by 0.8, 1.3, 2.2, and 2.6 points, respectively. These results demonstrate the strong and consistent English-to-many translation performance of MSRT-4B.

\paragraph{English-centric vs. Many-to-Many Translation.}
Although optimized for many-to-many translation, MSRT-4B also performs strongly on English-source directions, achieving an average COMET score of 87.9 on {eng}$\rightarrow$44 and an overall score of 83.3 in Table~\ref{comet_1144}. These results suggest that future speech translation models should emphasize comprehensive multilingual optimization rather than focusing predominantly on English-centric directions.

\paragraph{High-Resource vs.~Low-Resource Languages.}
Table~\ref{tab:eng44_comet} shows that Qwen3-Omni performs strongly on high-resource languages such as Chinese (cmn) and Japanese (jpn), but degrades markedly on low-resource languages such as Burmese (mya) and Khmer (khm). In comparison, MSRT-4B maintains consistently strong performance across both resource regimes. This advantage stems from MoSE, whose trainable expert specializes in medium- and low-resource languages, while its frozen expert preserves high-resource performance.

\begin{table}[!t]
\centering
\renewcommand{\arraystretch}{0.95}
\setlength{\tabcolsep}{4pt}
\caption{COMET results on English $\rightarrow44$ directions.}
\label{tab:eng44_comet}
\begin{tabular}{lccccc}
\toprule
\multirow{2}{*}{\textbf{ISO}} & Gemini-3.5 & Seamless & Qwen3 & MCAT & MSRT \\
 & -Flash-Lite & M4T & -Omni & -27B & -4B \\
\midrule
ara & 83.8 & 84.5 & \textcolor{cvpr_blue}{\textbf{86.6}} & \underline{86.1} & 86.0 \\
azj & \underline{85.8} & 83.8 & 83.1 & 84.8 & \textcolor{cvpr_blue}{\textbf{87.1}} \\
ben & 85.1 & 84.6 & 83.3 & \underline{85.2} & \textcolor{cvpr_blue}{\textbf{86.2}} \\
bul & 88.9 & 88.8 & 89.1 & \underline{90.0} & \textcolor{cvpr_blue}{\textbf{90.3}} \\
cat & 85.3 & 85.0 & \underline{86.2} & 85.9 & \textcolor{cvpr_blue}{\textbf{86.9}} \\
ces & 89.1 & 88.0 & 88.9 & \underline{90.1} & \textcolor{cvpr_blue}{\textbf{90.4}} \\
cmn & 86.4 & 79.7 & \textcolor{cvpr_blue}{\textbf{88.3}} & 87.2 & \underline{87.8} \\
dan & 88.7 & 88.7 & 89.1 & \underline{89.5} & \textcolor{cvpr_blue}{\textbf{90.1}} \\
deu & 86.2 & 84.9 & \underline{86.8} & 86.5 & \textcolor{cvpr_blue}{\textbf{86.9}} \\
ell & 87.1 & 87.5 & 87.1 & \underline{88.7} & \textcolor{cvpr_blue}{\textbf{89.1}} \\
fas & 86.5 & 84.6 & 85.6 & \underline{86.9} & \textcolor{cvpr_blue}{\textbf{87.1}} \\
fin & 90.9 & 88.5 & 88.7 & \underline{91.3} & \textcolor{cvpr_blue}{\textbf{91.5}} \\
fra & 85.8 & 85.3 & \textcolor{cvpr_blue}{\textbf{87.4}} & 86.3 & \underline{86.8} \\
heb & 85.2 & 84.5 & 73.8 & \underline{87.2} & \textcolor{cvpr_blue}{\textbf{87.8}} \\
hin & \underline{79.0} & 78.1 & 77.7 & 78.9 & \textcolor{cvpr_blue}{\textbf{79.3}} \\
hrv & 89.0 & 87.8 & 87.8 & \underline{89.2} & \textcolor{cvpr_blue}{\textbf{90.0}} \\
hun & \underline{88.1} & 86.0 & 86.8 & 87.6 & \textcolor{cvpr_blue}{\textbf{88.4}} \\
ind & 89.8 & 89.0 & \textcolor{cvpr_blue}{\textbf{91.0}} & 90.2 & \underline{90.7} \\
ita & 86.3 & 85.1 & \textcolor{cvpr_blue}{\textbf{87.3}} & \underline{86.8} & \textcolor{cvpr_blue}{\textbf{87.3}} \\
jpn & 89.7 & 84.7 & \textcolor{cvpr_blue}{\textbf{90.9}} & 90.4 & \underline{90.6} \\
kaz & \underline{88.6} & 87.9 & 85.5 & 88.2 & \textcolor{cvpr_blue}{\textbf{89.9}} \\
khm & \underline{81.4} & 79.9 & 77.4 & 79.7 & \textcolor{cvpr_blue}{\textbf{83.9}} \\
kor & 88.4 & 85.1 & \textcolor{cvpr_blue}{\textbf{89.7}} & 88.5 & \underline{89.0} \\
lao & 81.4 & 81.4 & 80.3 & \underline{83.1} & \textcolor{cvpr_blue}{\textbf{84.5}} \\
msa & 87.6 & 86.6 & \underline{88.2} & 87.5 & \textcolor{cvpr_blue}{\textbf{88.5}} \\
mya & \underline{87.1} & 85.7 & 71.9 & 85.0 & \textcolor{cvpr_blue}{\textbf{88.3}} \\
nld & 85.5 & 85.1 & 86.0 & \underline{86.6} & \textcolor{cvpr_blue}{\textbf{87.0}} \\
nob & 87.9 & 86.8 & 88.4 & \underline{88.7} & \textcolor{cvpr_blue}{\textbf{89.2}} \\
pol & 87.1 & 85.7 & 87.3 & \underline{88.1} & \textcolor{cvpr_blue}{\textbf{88.3}} \\
por & 87.3 & 86.6 & \textcolor{cvpr_blue}{\textbf{88.5}} & 87.9 & \underline{88.2} \\
ron & 88.5 & 87.8 & 88.8 & \underline{89.1} & \textcolor{cvpr_blue}{\textbf{89.8}} \\
rus & 87.8 & 86.3 & \textcolor{cvpr_blue}{\textbf{88.9}} & \underline{88.8} & \underline{88.8} \\
slk & 88.4 & 87.3 & 87.1 & \underline{88.8} & \textcolor{cvpr_blue}{\textbf{89.6}} \\
slv & 87.8 & 86.8 & 85.5 & \underline{88.5} & \textcolor{cvpr_blue}{\textbf{89.0}} \\
spa & 84.2 & 83.2 & \underline{85.4} & 85.2 & \textcolor{cvpr_blue}{\textbf{85.5}} \\
swe & 88.7 & 88.4 & 88.7 & \underline{89.3} & \textcolor{cvpr_blue}{\textbf{89.9}} \\
tam & 87.6 & 87.3 & 85.3 & \underline{88.3} & \textcolor{cvpr_blue}{\textbf{88.5}} \\
tha & 87.1 & 82.1 & 80.1 & \underline{83.3} & \textcolor{cvpr_blue}{\textbf{88.3}} \\
tgl & 82.9 & 83.3 & \textcolor{cvpr_blue}{\textbf{88.9}} & \underline{87.7} & 83.9 \\
tur & \underline{88.3} & 86.7 & 88.1 & 88.2 & \textcolor{cvpr_blue}{\textbf{88.7}} \\
urd & \underline{81.2} & 79.4 & 78.3 & 80.9 & \textcolor{cvpr_blue}{\textbf{82.0}} \\
uzb & 88.2 & 87.7 & 79.7 & \underline{88.2} & \textcolor{cvpr_blue}{\textbf{89.6}} \\
vie & 87.1 & 85.6 & \textcolor{cvpr_blue}{\textbf{88.6}} & 87.8 & \underline{88.3} \\
yue & 81.7 & 79.8 & \textcolor{cvpr_blue}{\textbf{88.6}} & 88.0 & \underline{88.3} \\
\midrule
\textbf{Avg.} & 86.6 & 85.3 & 85.7 & \underline{87.1} & \textcolor{cvpr_blue}{\textbf{87.9}} \\
\bottomrule
\end{tabular}
\end{table}

\begin{table}[h]
\centering
\caption{COMET results across the $45\times44$ directions.}
\label{tab:model_score_distribution}
\begin{tabular}{lrrr}
\toprule
\textbf{Model} & \textbf{$\geq80$} & \textbf{$[70,80)$} & \textbf{$<70$} \\
\midrule
Whisper+NLLB-3.3B & 1038 & 680 & 262 \\
Gemini-3.5-Flash-Lite & 928 & 815 & 237 \\
SeamlessM4T-V2-Large & 117 & 1109 & 754 \\
Qwen3-Omni-30B-Instruct & 881 & 557 & 542 \\
MCAT-27B & 1232 & 554 & 194 \\
MSRT-4B (ours) & \textbf{1552} & 359 & 69 \\
\bottomrule
\end{tabular}
\end{table}

\begin{figure}[h]
\centering
 \includegraphics[width=\linewidth]{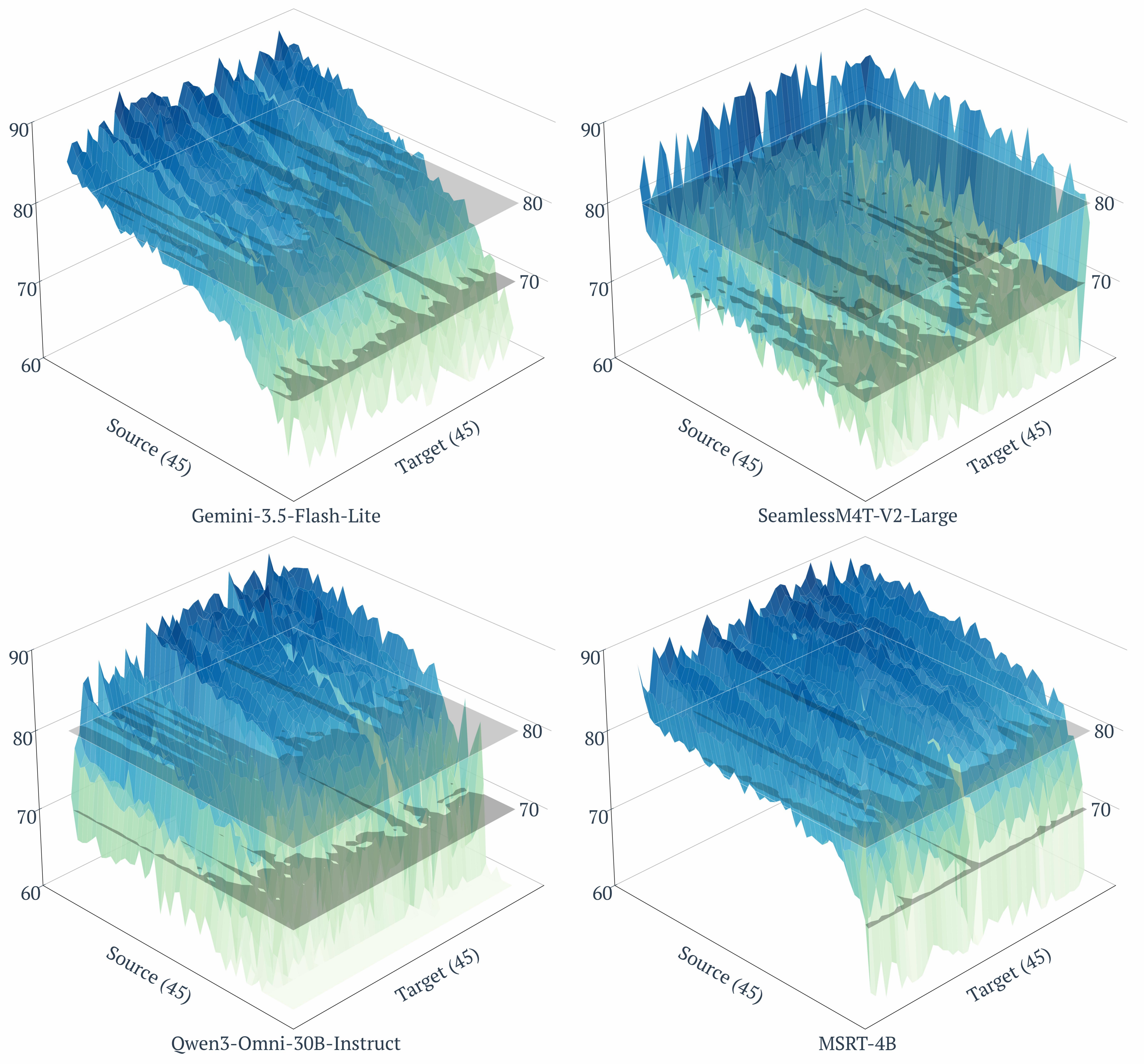}
\caption{COMET results for all directions. Shaded regions highlight scores falling below 80 and 70. Scores for identical source and target languages, such as eng$\rightarrow$eng along the diagonal, are smoothed for visualization.}
\label{fig:4545}
\end{figure}

\subsection{Systematic Analysis on $45\times44$ Directions}

\paragraph{Main Results.}
Table~\ref{tab:model_score_distribution} shows that MSRT-4B achieves COMET scores of at least 80 on 1,552 of the 1,980 directions. MCAT-27B, Whisper+NLLB-3.3B, Gemini-3.5-Flash-Lite, Qwen3-Omni-30B-Instruct, and SeamlessM4T-V2-Large achieve scores of at least 80 on 1,232, 1,038, 928, 881, and 117 directions, respectively. These results demonstrate the broader and more reliable many-to-many translation capabilities of MSRT-4B. 

\paragraph{API Comparison.}
MSRT-4B outperforms the Gemini-3.5-Flash-Lite API, achieving COMET scores of at least 80 on 1,552 directions, compared with 928 for Gemini, and scores below 70 on only 69 directions, compared with 237 for Gemini. This result demonstrates broader language coverage and stronger overall performance than the API.

\paragraph{English-Centric Bias.}
SeamlessM4T-V2-Large is centered on English, with its strongest performance concentrated on directions involving English. Its quality drops on non-English-to-non-English translation, yielding only 117 directions with COMET scores of at least 80 and 754 with scores below 70. In contrast, MSRT-4B remains consistent across the translation directions.

\paragraph{Parameter Efficiency.}
With only 4B parameters, MSRT-4B surpasses the much larger Qwen3-Omni-30B-Instruct and MCAT-27B, achieving COMET scores of at least 80 on 1,552 directions versus 881 and 1,232, respectively. It also produces fewer low-scoring directions, demonstrating efficient use of model capacity and broader multilingual coverage without relying on model scale.

\begin{figure*}[!t]
 \centering 
    \includegraphics[width=\linewidth]{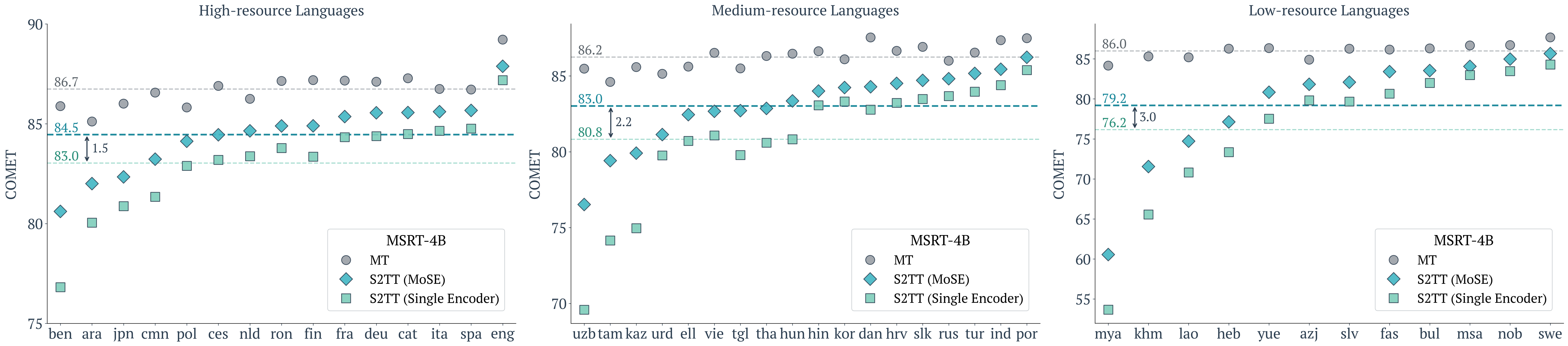}
\caption{MoSE ablation and S2TT--MT comparison across resource levels. Consistent gains, largest for low-resource languages, show that MoSE breaks the curse of multilinguality.}
    \label{fig:mse}
\end{figure*}

\begin{table}[t]
\centering
\setlength{\tabcolsep}{2.5pt}
\caption{Ablation results of MSRT-4B on $X\!\rightarrow\!44$ translation.}
\label{tab:ablation_english}
\small
\begin{tabular}{lccccccc}
\toprule
\textbf{Configuration}
& \textbf{ara}
& \textbf{ben}
& \textbf{khm}
& \textbf{mya}
& \textbf{tam}
& \textbf{uzb}
& \textbf{Avg.} \\
\midrule
MSRT-4B (2 experts)
& 82.0 & 80.6 & 71.5 & 60.5 & 79.4 & 76.5 & 75.1 \\
\quad 3 experts
& 81.9 & 80.7 & 71.5 & 62.5 & 78.9 & 76.6 & 75.3  \\
\quad 1 expert
& 80.0 & 76.8 & 65.6 & 53.7 & 74.2 & 69.6 & 70.0  \\
\quad w/o LoRA
& 81.2 & 78.9 & 68.7 & 56.6 & 77.7 & 75.5 & 73.1  \\
\bottomrule
\end{tabular}
\end{table}

\subsection{Ablation Study}

\paragraph{Component Ablations.}
Table~\ref{tab:ablation_english} evaluates MSRT-4B on six relatively low-performing source languages, with two selected from each resource group, translated into the other 44 languages.
\begin{itemize}
    \item \textbf{Number of experts.} Three-way routing, with one expert for each resource group, performs similarly to two-way routing between high- and medium/low-resource languages (75.3 vs.~75.1), whereas using a single expert substantially reduces the average COMET score to 70.0. These results show that two experts provide sufficient resource-aware specialization for the current 45 languages, while larger language inventories may benefit from finer-grained expert partitioning.
    \item \textbf{LoRA.} Removing LoRA lowers average COMET from 75.1 to 73.1, showing that lightweight decoder adaptation complements MoSE by mapping speech representations into the LLM's text space.
\end{itemize}

\paragraph{Resource-Level Analysis.}
Figure~\ref{fig:mse} compares MoSE variants and text MT across 45 languages by resource level; each score averages the other 44 target directions.
\begin{itemize}
    \item \textbf{MoSE gains across 45 languages.} MoSE improves COMET by 3.0, 2.2, and 1.4 points for low-, medium-, and high-resource languages, respectively. Gains across all groups show that MoSE breaks the curse of multilinguality by substantially improving low-resource translation while also strengthening high-resource performance.
    \item \textbf{Text MT vs.~S2TT.} Text MT with source transcriptions provides an oracle upper bound for S2TT. Its gap to MSRT-4B narrows as speech resources increase, identifying speech representation and cross-modal alignment as the main bottlenecks.
\end{itemize}

\begin{figure}[t]
    \centering
    \includegraphics[width=\linewidth]{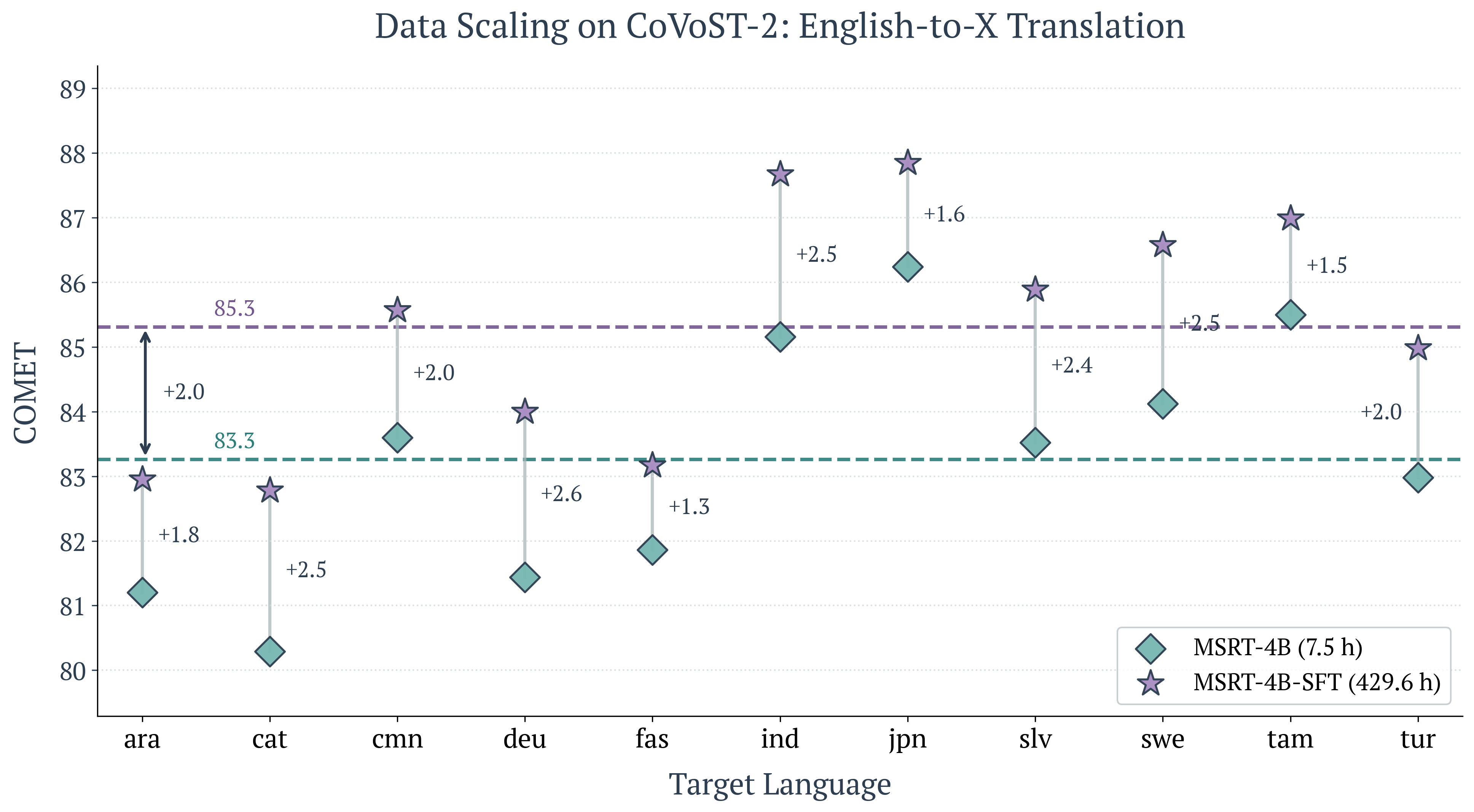}
    \caption{Data scaling on 11 CoVoST-2 English-to-X directions. MSRT-4B and MSRT-4B-SFT use 7.5 hours of FLEURS and 429.6 hours of CoVoST-2 speech, respectively; dashed lines denote average COMET.}
    \label{fig:covost2_scaling}
\end{figure}

\paragraph{Data Scaling.}
Figure~\ref{fig:covost2_scaling} shows strong zero-shot transfer: with only \textbf{7.5\,h} of FLEURS English speech, MSRT achieves scores of at least 80 on all 11 English-to-X directions, averaging 83.3. Scaling to \textbf{429.6\,h} of CoVoST-2 speech ($57.3\times$) raises the average to \textbf{85.3}, with consistent gains of 1.3--2.6 points across directions.

\section{Conclusion}

We introduced MSRT, a 4B many-to-many S2TT framework combining resource-aware MoSE with five-stage curriculum learning. By separating high-resource preservation from medium- and low-resource adaptation, MoSE improves all resource groups, with the largest gains on low-resource speech. Evaluation of all 1,980 directions among 45 languages demonstrates strong multilingual consistency, parameter efficiency, and cross-dataset generalization, showing that MoSE effectively breaks the curse of multilinguality. Future work will extend MoSE to broader language coverage.

\section*{Limitations}
MSRT derives its S2TT capability from the machine translation knowledge of the pretrained LLM. Consequently, its translation quality is bounded by the LLM's underlying MT performance, particularly for languages and directions that are weakly represented during pretraining.

\bibliography{aaai2027}

\end{document}